\documentclass[10pt,journal,compsoc]{IEEEtran}
\usepackage{cite}
\usepackage{tabularx}
\usepackage{amsmath,amssymb,amsfonts}
\usepackage{graphicx}
\usepackage{textcomp}

\usepackage{textcomp}
\usepackage{xcolor}
\usepackage{dsfont}
\usepackage{bm}
\usepackage{hyperref}

\usepackage[left=0.625in, right=0.625in, top=0.7in, bottom=1.0in]{geometry}

\usepackage{subcaption}
\usepackage{algorithm}

\usepackage{mathtools}
\usepackage{optidef}
\usepackage{amsthm}
\usepackage{color}

\usepackage{algorithm}
\usepackage{algpseudocode}
\usepackage{caption}
\usepackage{subcaption}
\usepackage{verbatim}
\usepackage{booktabs}

\usepackage{dsfont}
\usepackage{bm}

\usepackage{tikz}

\newcommand\copyrighttext{%

  \footnotesize \textcopyright \the\year{} IEEE. Personal use of this material is permitted.  Permission from IEEE must be obtained for all other uses, in any current or future media, including reprinting/republishing this material for advertising or promotional purposes, creating new collective works, for resale or redistribution to servers or lists, or reuse of any copyrighted component of this work in other works.}

\newcommand\copyrightnotice{%
\begin{tikzpicture}[remember picture,overlay]
\node[anchor=south,yshift=10pt] at (current page.south) {\fbox{\parbox{\dimexpr0.75\textwidth-\fboxsep-\fboxrule\relax}{\copyrighttext}}};
\end{tikzpicture}%
}

\begin{document}

\title{Exploring the Privacy-Energy Consumption Tradeoff for Split Federated Learning}
\author{Joohyung~Lee,~\IEEEmembership{Senior Member,~IEEE,} Mohamed Seif, Jungchan Cho,~\IEEEmembership{Member,~IEEE,}\\and H. Vincent Poor,~\IEEEmembership{Life Fellow,~IEEE}



\copyrightnotice
\thanks{Corresponding author: Jungchan Cho.}}

\markboth{Journal of \LaTeX\ Class Files,~Vol.~14, No.~8, August~2021}%
{Shell \MakeLowercase{\textit{et al.}}: A Sample Article Using IEEEtran.cls for IEEE Journals}

\IEEEpubid{0000--0000/00\$00.00~\copyright~2021 IEEE}

\maketitle

\begin{abstract}
Split Federated Learning (SFL) has recently emerged as a promising distributed learning technology that leverages the strengths of both federated and split learning. In this approach, clients are responsible for training only part of the model, termed the client-side model, thereby alleviating their computational burden. Then, clients can enhance their convergence speed by synchronizing these client-side models. Consequently, SFL has received significant attention from both industry and academia, with diverse applications in 6G networks. However, while offering considerable benefits, SFL introduces additional communication overhead when interacting with servers. Moreover, the current SFL method presents several privacy concerns during frequent interactions. In this context, the choice of the cut layer in SFL, which splits the model into both client- and server-side models, can substantially impact the energy consumption of clients and their privacy because it influences the training burden and output of the client-side models. Correspondingly, extensive research is required to analyze the impact of cut layer selection, and careful consideration should be given to this aspect. Therefore, this study provides a comprehensive overview of the SFL process and reviews its state-of-the-art. We thoroughly analyze the energy consumption and privacy regarding the cut layer selection in SFL by considering the influence of various system parameters on the selection strategy. Moreover, we provide an illustrative example of cut layer selection to minimize the clients' risk of reconstructing raw data at the server. This is done while sustaining energy consumption within the required energy budget, which involves trade-offs. We also discuss other control variables that can be optimized in conjunction with the cut layer selection. Finally, we highlight open challenges in this field. These are promising avenues for future research and development.
\end{abstract}

\section{Introduction}
\IEEEPARstart{F}{ederated} Learning (FL) fundamentally addresses the challenges associated with centralized learning by distributing the training process across multiple clients, enabling parallel processing. This approach also helps to safeguard the privacy of raw data stored on clients by exchanging only the model parameters. However, FL requires local training for each client, which can significantly burden clients with limited battery power and computational resources when dealing with large models such as Deep Learning (DL). Split Learning (SL) has emerged as a solution to mitigate this problem. SL involves breaking down a full DL model into two sub-models that can be trained both on a main server and across distributed clients. This approach alleviates the local training burden associated with FL while preserving data privacy. Nevertheless, SL introduces its own set of challenges, primarily related to the training time overhead, owing to its relay-based training method. In this relay-based approach, only one client trains with the main server at any given time, whereas the other clients remain idle. This sequential training method leads to inefficient distributed processing and a long training latency. To address this challenge, various strategies have been proposed to parallelize the SL training process \cite{Wu2023}. Inspired by these efforts, split federated learning, simply called split-fed learning (SFL), has recently been proposed as a novel approach that leverages the strengths of both FL and SL. Unlike SL, in SFL, all clients perform their local training in parallel while actively engaging with the main server and federated server (fed server). In SFL, the fed server plays a pivotal role in aggregating local model updates from clients using predefined aggregation techniques, such as FedAvg. This aggregation process occurred synchronously during each round of training. By introducing this additional aggregation server, SFL seamlessly combines the advantages of both FL and SL \cite{Thapa2022}. 

 \begin{figure*}       
	\centering	\includegraphics[width=520pt,height=250pt]{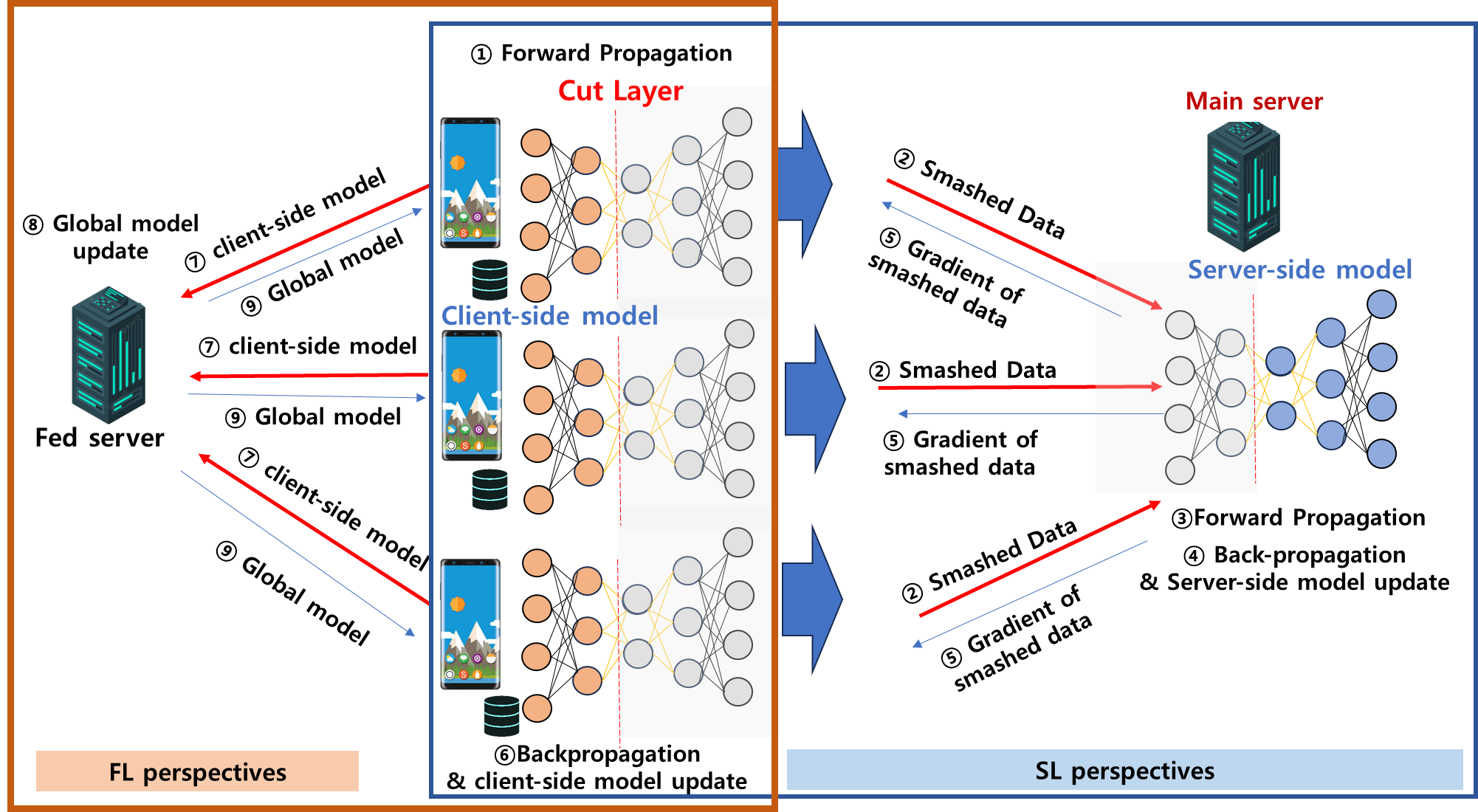}
    \caption{\small{Workflow of SFL: In FL, the full model undergoes local training rather than just a subset. In SL, training processes do not occur in parallel. SFL seamlessly integrates the benefits of both FL and SL by addressing their respective limitations.}}
    \label{fig:Figure_sfl_framework}
   \end{figure*}
   
Despite the advantageous integration of SL and FL in SFL, current SFL still presents several privacy concerns. Notably, the exchanged outputs of client-side models, known as smashed data, and the model updates between clients and servers are correlated with the raw data, leaving them susceptible to potential reconstruction attacks, where adversaries attempt to reconstruct the raw data from the correlated information. To address this growing concern, this study advocates the determination of an appropriate split point between client- and server-side models, specifically known as cut layer selection in SFL, to mitigate the risk of reconstruction attacks while sustaining energy consumption within the required energy budget. This is particularly important for clients with limited battery power. It is worth noting that the selection of the cut layer significantly influences the outcome of the smashed data, impacting the computational and communication overhead as well as the level of privacy. For example, as discussed in \cite{Duan2022}, an ongoing study focused on the adaptive selection of the cut layer, considering the varying computing and networking capabilities of clients. Nevertheless, prior research has not thoroughly analyzed both energy consumption and privacy in the context of cut layer selection, despite some studies focusing on empirical and analytical studies of reducing the overall latency in SFL. This motivated us to provide an analysis of both the energy consumption and privacy levels related to cut layer selection and a potential solution for developing novel cut layer selection techniques. 

The contributions of this article can be summarized as follows:
\begin{itemize}
    \item We provide a comprehensive overview of the overall process of SFL along with an in-depth examination of how the choice of the cut layer impacts SFL in terms of both privacy and energy consumption.
    \item Building upon this analysis, we propose a potential solution for optimizing cut layer selection, seeking to minimize the risk of reconstruction attacks while ensuring that energy consumption remains within the specified energy budget.
    \item Finally, we shed light on several prospective avenues for future research and conclude the paper.
\end{itemize}

\section{Why is Cut Layer Selection Important?}
\subsection{Background about SFL}
The SFL framework consists of two servers: i) a fed server and ii) a main server with multiple clients. The entire model is divided into two distinct sub-models: i) the client-side model and ii) the server-side model. The workflow of SFL is shown in Fig. ~\ref{fig:Figure_sfl_framework}. Each client performs forward propagation on the client-side model using its dataset, passing the smashed data and corresponding labels to the centralized main server (Step 1-2). Notably, the smashed data are features extracted from the dataset via forward propagation using a client-side model. The main server performs forward propagation from the smashed data on the server-side model and backpropagation by calculating the loss between the true and predicted labels, which can be performed in parallel. Subsequently, the server-side model is obtained using FedAvg (Step 3-4) and the main server passes the gradient of the smashed data to each client for the client-side local model update (Step 5-6). The fed server then receives all the clients' updated client-side local models and aggregates them using FedAvg (Step 7-8). Finally, the fed server sends all clients the updated client-side global model, enabling the synchronization of the client-side models (Step 9). In summary, in the SFL approach, clients are responsible for training only a part of the model, which reduces their computational burden compared with traditional FL. Furthermore, SFL enhances the convergence speed of SL by synchronizing the client-side models through the fed server. Nonetheless, interacting with two servers in SFL introduces additional communication overheads.

   \begin{figure*}       
	    \centering	\includegraphics[width=520pt,height=220pt]{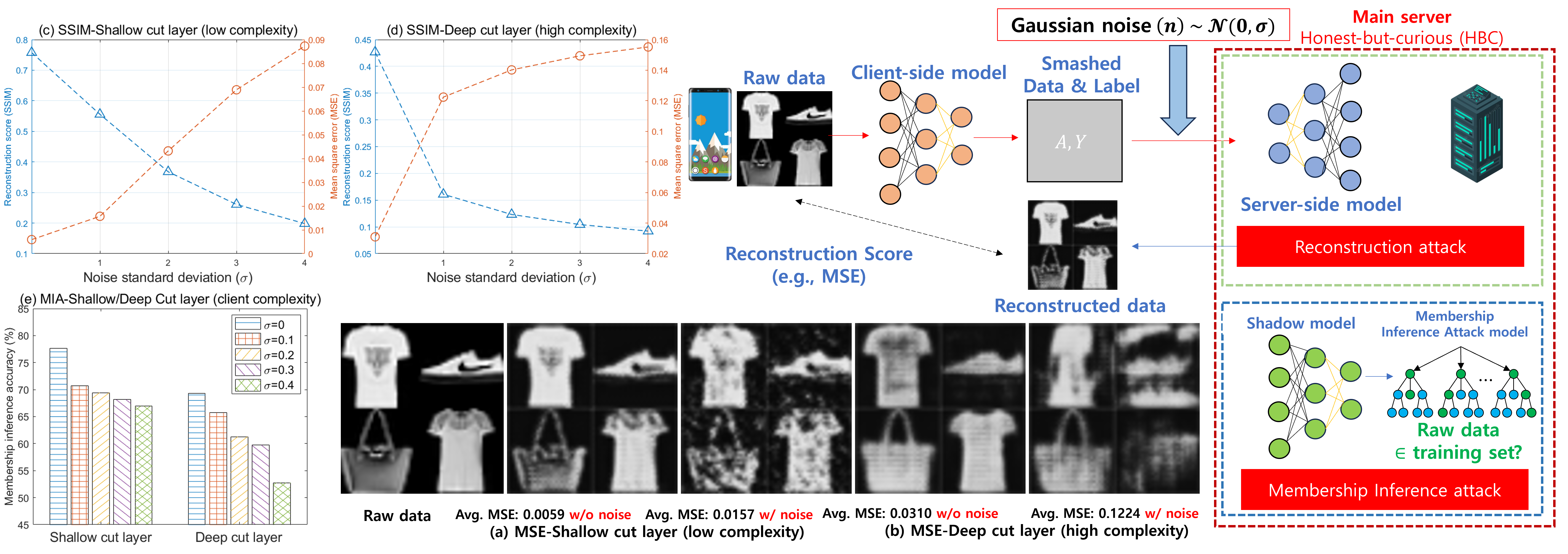}
        \caption{\small{Example of reconstruction and membership inference attacks in SFL as a privacy concern: This highlights the tradeoff between the client model complexity and privacy. Note that A reconstruction attack is an attempt to reconstruct the original input data from the smashed data. Additionally, membership inference attacks are another type of attack, aiming to determine whether a given data record was part of the target model’s training dataset or not (The decision tree-based attack model can be employed in such scenarios.) \cite{chen2020practical}.}}
    \label{fig:Figure_privacy_issues}
   \end{figure*}
   
\subsection{A Brief Survey of SFL}
As previously discussed, SFL imposes a significant communication overhead. This is primarily owing to the transmission of smashed data, gradients, and model updates between clients, the main server, and the fed server, indicating the need for improvements \cite{Yujia2023}. Nevertheless, there remains a scarcity of research focusing on the analytical modeling of the SFL, in contrast to the abundance of empirical studies \cite{Zhang2023}. Notably, \cite{Han2021,Thapa2022} developed analytical models that investigated the effect of the total model training time (i.e., latency) on the cut layer point. Correspondingly, based on such analyses, various approaches have been proposed to improve communication efficiency in SFL \cite{Yujia2023, Han2021, Wu2023}. Specifically, \cite{Han2021} introduced an optimal cut layer selection method based on latency analysis. Then, a convergence analysis of the SFL framework is conducted. \cite{Wu2023} extended the latency analysis of SFL considering wireless networks. They then designed a joint optimization problem involving cut layer selection, client clustering, and bandwidth allocation to minimize the overall training latency over wireless networks. Similarly, \cite{Zhu2024} proposed a joint cut layer selection and server computation resource allocation method for clients. \cite{Yujia2023} employed an auxiliary network for local updates of client-side models, maintaining only a single server-side model to reduce storage costs on the main server. They conducted a thorough analysis of communication costs, storage space, and convergence. The work of \cite{Lin2024} focused on reducing the dimensions of activations' gradients for backpropagation to reduce the communication overhead. Additionally, the study explored joint optimization strategies for subchannel allocation, power control, and cut layer selection in wireless networks, aimed at minimizing the latency per round.

Moreover, in SFL, privacy concerns arise from the interactions between clients and servers, including the main and fed servers. In the forward phase, when clients transmit smashed data to the main server, the data become susceptible to reconstruction attacks, jeopardizing the privacy of original information. Furthermore, the exchange of model updates with the fed server introduces another potential vulnerability to raw data. Several privacy-preserving mechanisms can be employed to mitigate these risks. In a recent study \cite{Kim2020} on SL, researchers conducted an empirical investigation of the impact of selecting a cut layer on reconstruction attacks during the forward phase. Their findings indicated that a greater depth of layers on the client side implied more non linear functions, which compressed the raw data by eliminating less informative features, thereby enhancing its resistance to reconstruction attacks. Similarly, \cite{Vepakomma2019} presented empirical studies on how the distance correlation between raw and smashed data influences privacy leakage in SL on the selection of the cut layer. 

Remarkably, most previous studies concentrated on analyzing and optimizing the latency in SFL for cut layer selection strategies. However, there is a notable absence of studies addressing the impact of energy consumption on SFL. Here, the choice of cut layer significantly affects the training burden and the outcomes of client-side models, directly influencing energy consumption. This is crucial for estimating the associated costs for clients with limited battery power. Furthermore, there has been a lack of exploration into the joint consideration of energy consumption and privacy levels in this domain. In this context, building on the insight of SL regarding privacy concerns in previous studies, we conduct empirical studies of SFL, which can be extended to explore cut layer selection strategies that strike a balance between energy consumption and privacy levels.

\begin{figure*}       
	\centering	\includegraphics[width=500pt, height=170pt]{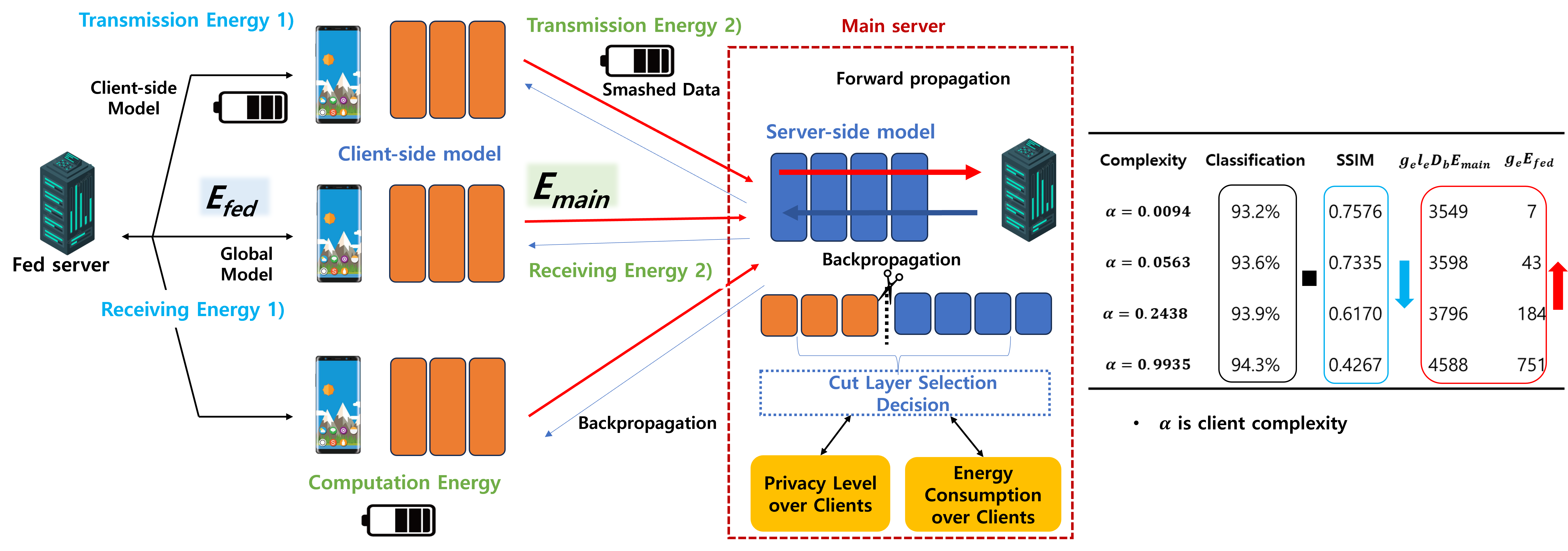}
    \caption{\small{Proposed cut layer selection considering both energy consumption and privacy level.}}    \label{fig:Figure_framework}
   \end{figure*}

\subsection{Cut Layer Selection Impact}
\subsubsection{Energy Efficiency Perspective}
In the context of SFL, client energy consumption occurs from both networking and computing perspectives. Specifically, from a computational perspective, clients engage in local training on their datasets using client-side models. From a networking perspective, there is energy expenditure associated with uplink transmission and downlink reception between clients, the federated server, and the main server. The downlink receiving energy consumption tends to be negligible because of the relatively small amount of energy required compared with the uplink transmission power. For uplink transmission, energy is required to transmit the smashed data to the main server. In addition, each client must transmit its client-side local model to the fed server for aggregation. Notably, the cut layer selection strategy can affect the size of the client-side model (i.e., model complexity). Consequently, as the size of the client-side model increases, the energy consumption for both the computation and model transmission to the fed server also increases. Therefore, to reduce energy consumption, a sophisticated management strategy for cut layer selection should be considered.

\subsubsection{Privacy Level Perspective}
As shown in Fig. \ref{fig:Figure_privacy_issues}, when we assume that the main server is honest but curious (HBC), it is still possible for the main server to reconstruct the original input data from the smashed data (specifically, the embedded features) received from clients. This process is known as a reconstruction attack. In this context, we can assess the privacy risks associated with this reconstruction attack by measuring the human perceptual similarity between the original and reconstructed images, known as the structural similarity index (SSIM)\footnote{This study concentrates on image data in the reconstruction attack, a prevalent focus in this area. Recently, the significance of text data privacy in this attack has also been recognized. In this case, metrics such as recovery rate, fluency, and attack accuracy are used to evaluate the privacy risk.}. The SSIM formula compares two images by measuring three factors: luminance, contrast, and structure. The comparison was performed as $\operatorname{SSIM}(x, y) = l(x, y) c(x, y) s(x, y)$, where $x$ and $y$ are the pixel values of the compared images. The luminance comparison is denoted by $l$, the contrast comparison by $c$, and the structural comparison by $s$. This metric, which ranges from [0, 1], is widely used to measure attack and defense performances \cite{Wu2023_2}, where the value $1$ denotes the most similar and vice versa. Figs. 2 (a)–(d) illustrate the relationship between the client model complexity and privacy leakage. This highlights the effects of altering the cut layer on the quality of the images reconstructed on the main server. Notably, as the cut layer index increased, the quality of the reconstructed image degraded, indicating an increase in model complexity. This degradation is interchangeable with enhanced privacy. This is because the complexity of the model deployed at the client end introduces advanced non linearities into the output. This added intricacy can make it more challenging for adversaries to reverse-engineer and retrieve private input data. Moreover, Fig. \ref{fig:Figure_privacy_issues} (a)-(d) also presents an illustrative example to practically demonstrate the application of noise addition, a fundamental strategy in achieving differential privacy. Here, we focus on a widely adopted noise addition mechanism, the \textit{Gaussian} mechanism, which is instrumental in introducing carefully calibrated noise to sensitive data, effectively masking individual information while preserving the utility of the overall dataset. The Gaussian mechanism employs a Gaussian distribution for this purpose, providing distinct privacy-accuracy trade-offs. In this context, it is evident that introducing Gaussian noise to the smashed data further degrades the quality of the reconstructed image. This can be indicated by the mean square error (MSE), representing the difference in pixels between the original and reconstructed images. As the level of Gaussian noise increased, the MSE also increased, leading to a decrease in the SSIM in both the shallow and deep cut layer scenarios. We also conducted a privacy assessment of the Membership Inference Attack (MIA), another type of attack on data privacy, to determine whether a given data record was part of the target model’s training dataset or not. The main server can still exploit these vulnerabilities to train a binary classifier and infer membership information by taking the predicted value of the data sample obtained by querying the target model as the input to the binary attack model. In this context, we can assess the privacy risks associated with this MIA by measuring accuracy. This is widely used as a metric in MIA: the percentage of data records correctly inferred by the attack model in the entire given dataset. Transfer-inherited shadow learning is used to train the attack model \cite{chen2020practical}. Fig. \ref{fig:Figure_privacy_issues} (e) illustrates that MIA accuracy decreases with deeper cut layers and higher noise levels. Cutting in a shallow layer results in the main server retaining significant information regarding the entire target model, requiring less effort to create a shadow client model. This makes the SFL vulnerable to MIA attacks, whereas the attack accuracy becomes more resilient to noise.

\section{Case Study: Cut Layer Selection}
\subsection{An Illustrative Example}
Based on the latency analysis of SFL in \cite{Thapa2022} and \cite{Han2021}, and leveraging the energy consumption insights gained from FL studies \cite{Lee2024}, this section examines the overall energy consumption associated with SFL. The analysis was based on the vanilla version of SFL rather than SFLv2/v3. However, core concepts and intuition are easily applicable to higher versions. Similar to the FL, as shown in Fig. \ref{fig:Figure_framework}, the total energy consumption $E$ of each client during SFL encompasses both i) computing and ii) networking energy. However, it requires customization to account for model splitting, interactions with both the fed and main servers, and distinctive features introduced in the SFL. Taking these considerations into account, to formulate the energy consumption model, consider a scenario with $K$ clients. Subsequently, $K$ clients engage in the SFL framework for $g_e$ global epochs, which represent the overall number of rounds necessary to achieve a specific training loss. The client-side models were synchronized at each global epoch, enabled by the fed server. In each global epoch, by leveraging this synchronized client-side model, both the clients and main server conduct training for $l_e$ local iterations. Specifically, for simplicity, as in \cite{Thapa2022} and \cite{Han2021}, by considering homogeneous clients in a single local iteration, each client trains $D_b$ randomly sampled data items, commonly referred to as minibatches, from its dataset. By using the transmission power $P_t$, they can communicate with the fed server and main server with the total uplink (downlink) transmission rates $R_1^U$ and $R_2^U$ ($R_1^D$ and $R_2^D$), respectively. Consequently, the uplink (downlink) transmission rates for each client are reduced to $\frac{R_1^U}{K}$, $\frac{R_2^U}{K}$ ($\frac{R_1^D}{K}$ and $\frac{R_2^D}{K}$), respectively. Note that the entire model $W = [W_C;W_S],$ where $W_C$ and $W_S$ are the client- and server-side models, respectively. As in \cite{Thapa2022} and \cite{Han2021}, we assumed the same size of smashed data, denoted as $q$ sent from the clients to the main server. Let $|W|$ be the number of model parameters in the entire model $W$ where all model parameters have the same size $b$, and let $\alpha$ be the fraction of model parameters that serves as the model cut layer point for SFL in $W_C$, where $|W_C| = \alpha|W|$ and $|W_S| = (1-\alpha)|W|$\footnote{As the depth of a DL model increases, there is a corresponding rise in both the number of learnable parameters and multiply-add floating-point operations. Conversely, SFL divides the layers of a model and assigns only the fraction determined by $\alpha$ to a client, indicating that lower $\alpha$ values correspond to reduced client-side model complexity.}.
In addition, $T$ represents the time required for forward and backward propagations in the full model. In backward propagation, gradients of a constant size, denoted by $q'$, are transmitted. Here, $P_c$ represents the power consumption necessary to train the full model $W$ at the client level, which is calculated as the total energy consumption required to train the entire model divided by the duration $T$. Therefore, in SFL, energy consumption during interaction with the main server, denoted as $E_{\text{main}}$, requires each client i) to proceed forward and backward propagation using $W_C$, incurring energy consumption $\alpha TP_c$ and ii) to transmit the smashed data (or receive the gradients) to (from) the main server per each sampled data item, incurring energy consumption $\frac{qK}{R_2^U}P_t$ (or $\frac{q'K}{R_2^D}P_r$), where $P_r$ is the receiving power consumption of each client. In parallel, the energy consumption for interaction with the fed server, denoted as $E_{\text{fed}}$, requires each client to i) send its local model to the fed server aggregation, incurring energy consumption $\frac{\alpha b|W|K}{R_1^U}P_t$, and ii) receive the updated global model from the fed server, incurring energy consumption $\frac{\alpha b|W|K}{R_1^D}P_r$ during one global epoch. Finally, the total energy consumption $E(\alpha)$ of each client for the SFL is given by
\begin{align}
\label{eq:opt_prob}
E(\alpha)=g_e(l_eD_bE_{\text{main}}+E_{\text{fed}}).
\end{align}
As discussed previously, under the assumption of an HBC main server, the privacy risks associated with a reconstruction attack can be evaluated by measuring the SSIM between the original input and reconstructed images with respect to $\alpha$, denoted as $RS(\alpha)$. To evaluate the impact of cut layer selection in terms of $RS(\alpha)$ and $E(\alpha)$, we utilized a training-based adversarial inversion approach. The reconstruction model was implemented to reflect the inverted structure of the client-side model $W_C$ with transposed convolutional and Tanh activation layers. In our experiment, the overall classifier model $W$ for SFL was designed with four convolution blocks (Conv-BN-ReLU-Conv-BN-ReLU-Conv-BN-Max-ReLU) and one linear block. The linear block consisted of two fully connected layers and a Softmax function. The cut layer was selected between the blocks to divide the client- and server-side models. For performance evaluation, the Fashion-MNIST dataset was used \footnote{https://github.com/zalandoresearch/fashion-mnist}. The input image was resized to $32\times32$ and $D_b$ was set to 128. An Adam optimizer was used, and the learning rate was set to 0.0002. The betas for Adam were set to (0.5, 0.999). The classifier model for SFL was trained for up to 50 global and 75 local training iterations. The reconstruction model was trained for up to 50 epochs using the same optimizer and settings as the classifier model. The detailed parameters are summarized in Table \ref{table:parameter}.

\begin{table}[h!]
\caption{Parameter Settings.} 
\centering 
\begin{tabular}{c c}
\hline\hline 
Parameters & Values
\\ [0.5ex] 
\hline 
$K$ & 5\\ 
$W$ & 31,484,464\\
$D_b$ & 128 \\
$q$, $q'$ & 491,520 bits, 491,520 bits\\
$b$ & 32 bits \\
$T$ & 0.00055 seconds \\
($R_1^U, R_2^U$), ($R_1^D$, $R_2^D$) & 200 Mbit/s, 100 Mbit/s  \\
$P_c$, $P_t$, $P_r$ & 4W, 0.2W, 0.2W\\ [1ex] 
\hline 
\end{tabular}
\label{table:parameter} 
\end{table}

  \begin{figure}       
	\centering	\includegraphics[width=220pt,keepaspectratio]{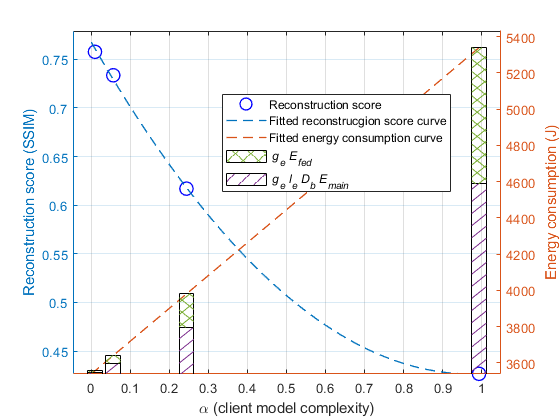}
    \caption{\small{Reconstruction score and energy consumption with respect to the depth of cut layer: This represents the tradeoff between energy consumption and privacy.}}
    \label{fig:Figure_graph1}
   \end{figure}

\begin{figure}       
	\centering	\includegraphics[width=220pt,keepaspectratio]{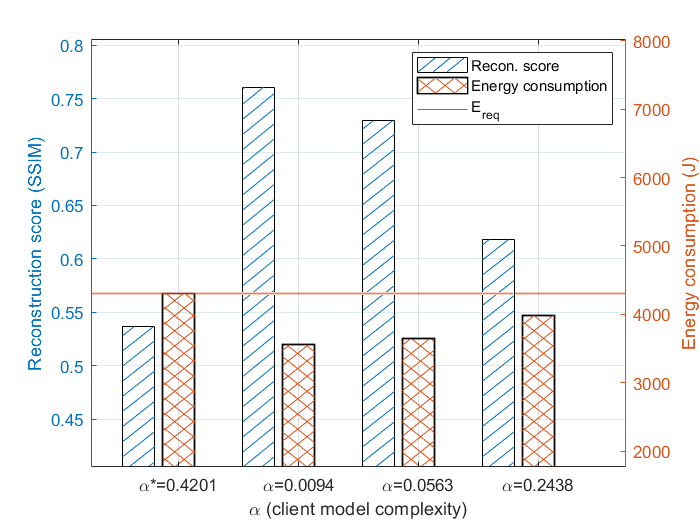}
    \caption{\small{Privacy gain of proposed scheme: It represents that optimal cut layer selection ($\alpha^*$=0.4201) achieves minimum privacy leakage (SSIM) while satisfying the required energy budget.}}
    \label{fig:Figure_graph2}
   \end{figure}

As shown in Fig. \ref{fig:Figure_graph1}, increasing the depth of the cut layer (i.e., increased complexity of the client-side model) leads to higher energy consumption ($E$), causing both $E_{\text{main}}$ and $E_{\text{fed}}$ to rise due to the increased computing and networking energy usage. Concurrently, RS consistently decreased with deeper cut layers. This phenomenon implies that the depth of the cut layer can be directly correlated with the complexity of the client-side model, potentially making attackers to reconstruct the image from the smashed data more difficult. Nevertheless, owing to the lack of an analytical understanding of RS, a data-driven approach can be applied to model RS through offline training using empirical measurements. In this context, we adopted regression-based modeling, which is one of the most widely used approaches, and includes mobile CPU property models (e.g., CPU power and temperature variation modeling). Consequently, we demonstrate that a convex model can effectively approximate the behavior of RS, yielding a Root Mean Square Error (RMSE) of only $0.0028$. Consequently, RS can be estimated as $\operatorname{RS}(\alpha)$ = $0.3597 \alpha^2 -0.7004 \alpha + 0.7675$. Note that the estimated $\operatorname{RS} (\alpha)$ may differ across applications; therefore, it should be pre-calculated offline before making any adjustments. Consequently, we can optimize the selection of the cut layer, denoted by $\alpha$, to minimize privacy leakage from reconstruction attacks, as measured by $RS(\alpha)$, while ensuring that energy consumption ($E$) remains within an acceptable energy budget ($E_{\text{req}}$). Therefore, the problem can be formulated as follows:
\begin{align}
\label{eq:opt_prob}
\begin{split}
&\underset{\alpha}{\text{minimize}} \quad \operatorname{RS}(\alpha) \hspace{0.1in}\text{s.t.}\quad E(\alpha) \leq E_{\text{req}},
\quad 0 \leq \alpha \leq 1.
\end{split}
\end{align}
Note that $E(\alpha)$ within the constraint should be calculated as the average across clients when dealing with heterogeneous client scenarios. The problem in (\ref{eq:opt_prob}) is a convex optimization problem because it features a convex objective function and affine constraints, which guarantee a global optimal solution, and can be solved simply using a CVX solver. Based on our empirical findings, the objective function $RS(\alpha)$ decreases strictly. Therefore, the optimal cut layer selection is expected to be located at the constraint boundary. Finally, assuming that each layer has a similar number of model parameters, $\alpha$ can be approximately mapped onto the index of the cut layer by computing $\left\lfloor \alpha \cdot \text{total number of layers} \right\rfloor$. Practically, this equation can be replaced with $\left\lfloor \alpha \cdot \text{total number of basic blocks} \right\rfloor$ $\cdot \text{number of layers per basic block}$, given that various DL models comprise basic blocks, with the entire network structure formed through the repeated linkage of these basic blocks. If the depth of the entire network is not excessively large, and the estimated $\operatorname{RS}(\alpha)$ and $E(\alpha)$ functions are not provided, an exhaustive search remains a viable option. However, if the depth is substantial, the search space can be narrowed by sampling a range of $\alpha$ values (e.g., $0.1, 0.2, 0.3, \cdots, 1$). This approach enables a more efficient and exhaustive search, leading to suboptimal solutions. Based on intuition from the analytical study, this exhaustive search involves verifying the energy budget constraint as the depth of the cut layer increases. Therefore, as illustrated in Fig. \ref{fig:Figure_graph2}, by optimizing the cut layer selection while ensuring the energy consumption budget, $\operatorname{RS}(\alpha)$ can be minimized in comparison to other cut layer selections within the feasible range of $\alpha$. This suggests that thoughtful cut layer selection is essential for optimizing the overall performance of SFL because it balances privacy levels with energy consumption. Furthermore, as shown in Fig. \ref{fig:Figure_framework}, it is essential to note that the selection of cut layers has minimal effect on model accuracy (e.g., 93.2\%-94.3\%) from our evaluation. This is because the choice primarily affects the distribution of the computational load between the client and the server without altering the integrity of the full model. This issue must be addressed using real-world examples. For example, the protection of personal information and maintenance of energy efficiency are crucial for supporting automated vehicle training. In addition, this can be critical in cases where Unmanned Aerial Vehicles (UAVs) with limited computing and battery power are owned by different service providers working together in the SFL \cite{Houda2023}.

\subsection{Other Types of Control Variables}
\begin{itemize}
    \item Computation/Power/Radio resource management: The computation resource, transmission power, and radio resource (i.e., bandwidth allocation over clients) of each client participating in SFL play a pivotal role in optimizing performance. However, due to the early stage of this field, there has been a lack of rigorous analysis for SFL. Fundamentally, those factors are known to affect overall latency, denoted as $T_i$ in our analysis, which in turn impacts energy consumption within the field of FL \cite{Lee2024}.   
   \item Client and dataset management: The high mobility of clients can affect the wireless network conditions between clients and the main server and the fed server, respectively. Additionally, variations in the size of datasets across clients, characterized by their unbalanced, non-independent, and identically distributed (non-IID) nature, may lead to slower convergence speeds. Given these challenges, implementing strategies for dynamic client selection and effective dataset management—such as downsampling or augmenting datasets—within SFL can be beneficial in improving both accuracy and convergence speeds.
\end{itemize}
As reviewed, the performance of SFL in terms of both energy efficiency and convergence speed is noteworthy. Heterogeneity in clients' computing and networking capabilities can affect the speed of convergence owing to the issue of stragglers, and this diversity also leads to differences in energy consumption among clients. Correspondingly, achieving an optimal balance among these multiple variables often requires substantial complexity when pursuing a global optimum. This necessitates the adoption of a modular design strategy or the employment of various heuristic methods to achieve suboptimal solutions.

\section{Open Challenges}
\subsection{Deep Reinforcement Learning}
Considering the dynamic and heterogeneous nature of client networking and computing capabilities, deep reinforcement learning (DRL) shows promise in intelligently selecting an optimal cut layer. This aims to balance privacy, energy consumption, and convergence speed in the SFL, along with other essential control parameters. A recent study by \cite{Xiao2023} illustrated how DRL is utilized to optimize cut layer selection for collaborative DL inference on multi-access computing (MEC) servers by efficiently offloading certain inference tasks from clients to the MEC server. The key challenges in applying DRL to address various SFL issues involve precisely defining the agents, environments, states, actions, and rewards. In addition, selecting an appropriate DRL model and fine-tuning hyperparameters become pivotal for improving convergence speed, particularly when dealing with complex and dynamic environments.

\subsection{Privacy and Security Protection}
As in \cite{Wu2023_2}, when clients and the main server hold their data and labels, respectively, attackers can be malicious data owners or eavesdroppers trying to intercept the gradients shared between the main server and clients. The similarity in cut layer gradients among data samples can reveal their connection to labels, which is known as a label-inference attack. Therefore, to mitigate risks, label-inference attacks must be addressed by carefully choosing the cut layer and implementing differential privacy (DP). Moreover, determining the cut layer is critical for maintaining security and integrity, particularly for model poisoning and data integrity. Specifically, the cut layer selection significantly influences a system's susceptibility to model-poisoning attacks. For instance, extensive client-side training increases the risk of client-side poisoning, complicating the defense against distributed attacks. Thus, sophisticated aggregation techniques are required to effectively counter poisoning risks. An inappropriately selected cut layer can also hinder data authenticity and integrity validation, posing challenges to maintaining the model's training integrity. Therefore, the selection of the cut layer requires careful consideration, balancing minimizing security risks and maximizing the efficiency of SFL.

\subsection {Lightweight Design}
To enhance the efficiency of SFL, it is beneficial to consider a quantization approach. In the context of SFL, adopting a quantization approach can substantially reduce the size of both local model updates and intermediate data, which is particularly useful for resource-constrained clients. This reduction has several advantages, including reduced bandwidth usage, lower transmission energy, and reduced storage requirements. Furthermore, quantization introduces an inherent privacy guarantee. This is because reducing the precision of the transmitted data inherently obfuscates exact values, thereby adding an extra layer of protection to sensitive information, which is crucial for maintaining data privacy. Moreover, to mitigate the burden of training on resource-constrained clients with reduced energy consumption, a strategy of downsizing the resolution and total volume of the training data can be employed, considering different data distributions and acceptable accuracy levels. In addition, the selection of a lightweight DL model or employing model-pruning techniques was adopted to facilitate a lightweight design for SFL.

\section{Conclusions}
This article provides an overview of SFL, focusing primarily on communication efficiency and privacy issues. By studying the impact of cut layer selection on both energy consumption and privacy, we provided a concrete example of efficient cut layer selection to minimize the risk of reconstruction attacks within the required energy budget. Finally, we suggest various other adjustable factors and highlight the promising research directions in this field.

\section*{Acknowledgments}
This work was supported by the U.S National Science Foundation under Grant ECCS-2335876.

\bibliographystyle{IEEEtran}

\section*{Biographies}
\vspace{-11pt}

\begin{IEEEbiographynophoto}{Joohyung Lee} (j17.lee@gachon.ac.kr) is currently an Associate Professor with the School of Computing, Gachon University and also a Visiting Fellow with the Department of Electrical and Computer Engineering, Princeton University. Before joining Gachon University, he was a Senior Engineer with Samsung Electronics From 2014 to 2017. His current research interests include federated learning, 6G networks, cloud/edge computing, smart grids, augmented reality, virtual reality, and network economics. 
\end{IEEEbiographynophoto}

\begin{IEEEbiographynophoto}{Mohamed Seif} (mseif@princeton.edu) is currently a Post-Doctoral Research Associate at Princeton University. His current research interests include information theory,
machine learning, and wireless communications.
\end{IEEEbiographynophoto}

\begin{IEEEbiographynophoto}{Jungchan Cho} (thinkai@gachon.ac.kr) is currently an Associate Professor in the School of Computing at Gachon University. From 2016 to 2019, he was a Senior Software Engineer at Samsung Electronics. His research interests include deep learning, computer vision, and machine learning.
\end{IEEEbiographynophoto}

\begin{IEEEbiographynophoto}{H. Vincent Poor} (poor@princeton.edu) is the Michael Henry Strater University Professor at Princeton University. His research interests are in the areas of information theory, machine learning and network science, and their applications in wireless networks, energy systems and related fields. Dr. Poor is a member of the U.S. National Academy of Engineering and the U.S. National Academy of Sciences. He received the IEEE Alexander Graham Bell Medal in 2017.
\end{IEEEbiographynophoto}

\vfill

\end{document}